\documentclass[runningheads]{llncs}
\usepackage[T1]{fontenc}
\usepackage{graphicx}
\usepackage{amsfonts}
\usepackage{amssymb}
\usepackage{todonotes}
\usepackage{amsmath}

\usepackage{orcidlink}
\usepackage[misc,geometry]{ifsym}

\usepackage{adjustbox}
\usepackage{subcaption}

\begin{document}
%
\title{Constraint-Reduced MILP with Local Outlier Factor Modeling for Plausible Counterfactual Explanations in Credit Approval}
\titlerunning{Constraint-Reduced MILP with Local Outlier Factor Modeling}
%

\author{Trung Nguyen Thanh\inst{1}\orcidlink{0009-0005-9847-7038} \and Huyen Giang Thi Thu\inst{2,3}\Letter\orcidlink{0009-0007-6283-3111} \and Tai Le Quy\inst{4}\orcidlink{0000-0001-8512-5854} \and Ha-Bang Ban\inst{1}\orcidlink{0000-0003-2241-5146} }

\authorrunning{T. Nguyen Thanh et al.}
%
\institute{Hanoi University of Science and Technology, Hanoi, Vietnam\\
\email{trungnguyenthanh.eworking@gmail.com}
\and Banking Academy of Vietnam, Hanoi, Vietnam\\ 
\and Vietnam Academy of Science and Technology, Hanoi, Vietnam\\
\email{huyengtt@hvnh.edu.vn}
\and University of Koblenz, Koblenz, Germany\\
\email{tailequy@uni-koblenz.de}
}

\maketitle        
\begin{abstract}
Counterfactual explanation (CE) is a widely used post-hoc method that provides individuals with actionable changes to alter an unfavorable prediction from a machine learning model. Plausible CE methods improve realism by considering data distribution characteristics, but their optimization models introduce a large number of constraints, leading to high computational cost. In this work, we revisit the DACE framework and propose a refined Mixed-Integer Linear Programming (MILP) formulation that significantly reduces the number of constraints in the local outlier factor (LOF) objective component. We also apply the method to a linear SVM classifier with standard scaler. The experimental results show that our approach achieves faster solving times while maintaining explanation quality. These results demonstrate the promise of more efficient LOF modeling in counterfactual explanation and data science applications.

\end{abstract}

\section{Introduction}
\label{sec:intro}
Evaluating the risk of a loan application is fundamental to the lending decision process of any financial institutions. Many credit scoring approaches have been introduced from statistical based to Artificial Intelligence (AI) based methods and ensemble classifiers \cite{dastile2020statistical}. Deep learning models, hybrid approaches, and ensemble classifiers have shown the ability to leverage multiple data sources and achieve higher classification accuracy compared to statistical models. However, these models are often perceived as ``black boxes'' by both decision-makers and loan applicants, which leads to challenges in providing explanations \cite{guidotti2018survey}.

With the rapid growth of financial services, many countries have also introduced stricter laws and regulations on accountability in credit scoring activities \cite{louzada2016classification}. Explainable AI (XAI) offers solutions to this challenge by seeking explanations for the decisions made by black-box models \cite{dessain2023cost,nallakaruppan2024explainable}. 
In particular, counterfactual explanation (CE) has emerged as a modern and effective approach in XAI \cite{stepin2021survey}. For example, a CE for a loan rejection decision can be formulated as follows \cite{wachter2017counterfactual}: ``the application was denied because the applicant's annual income was £30,000; had the income been £45,000, the loan would have been approved''.

A CE can be viewed as a search problem, with diverse research exploring different objectives and search methods such as optimization strategy, instance-based or heuristic search strategies \cite{guidotti2024counterfactual}. In which, DACE model~\cite{kanamori2020dace} is one of the earliest CE studies focusing on plausibility, using \textit{Mahalanobis distance}–based correlation analysis \cite{de2000mahalanobis} and \textit{local outlier factor}–based (LOF) \cite{breunig2000lof} risk measures. Then, \cite{cui2015optimal} and  \cite{ustun2019actionable} model the counterfactual search problem as integer programming or Mixed-Integer Linear Programming (MILP), using IBM ILOG CPLEX solver find optimal solutions. While offering high explainability, these approaches face limitations when the number of constraints grows. 
To address this problem, this paper introduces a method that reduces the number of constraints for LOF term from $\mathcal{O}(N^2)$ to $\mathcal{O}(N)$ in the MILP formulation. Moreover, we extend the DACE model on linear SVM classifier with the standard scaler.

The rest of our paper is structured as follows. Sect. \ref{sec:related} gives an overview of the related work. The technical background and our proposed methods are presented in Sect. \ref{sec:preliminary} and \ref{sec:method}, respectively. Sect. \ref{sec:evaluation} describes our experiments on various datasets. Finally, the conclusion and outlook are summarized in Sect. \ref{sec:conclusion}.

\section{Related work}
\label{sec:related}
MILP has been proposed as a means to obtain optimal counterfactual explanations. One of the earliest studies is \cite{cui2015optimal}. This work introduced logical representations and constraints to model the decision-making process of additive tree models, including random forest (RF), Adaboost, and Gradient boosting. The RF modeling in DACE \cite{kanamori2020dace} represents a specific instance of the logical approach presented in \cite{cui2015optimal}, but with different partitions. Subsequently, the work \cite{ustun2019actionable} also employed an MILP framework to obtain optimal solutions and compared it with DACE. Unlike DACE, this study focused on explaining the decisions of linear models through mixed constraints, with a cost function aimed at minimizing changes in quantiles under counterfactual actions. In addition, the study emphasized the feasibility of counterfactual actions by encoding conditional classification constraints for specific cases.
Other studies, such as \cite{romashov2022baycon}, demonstrated positive effects of integrating LOF into the objective function to enhance the plausibility of counterfactual actions. However, study \cite{nvemevcekgenerating} pointed out that although LOF serves as a meaningful penalty component in DACE, the linear constraint formulation for 1-LOF has a complexity of $\mathcal{O}(N^2)$, which poses a significant obstacle to scalability and further improvements in integer programming models.

\section{Preliminaries}
\label{sec:preliminary}

\subsection{Notations and Settings}

We denote ${1,2,\dots,n}$ by $[n]$ and use $\mathbb{I}(\psi)$ as the indicator of whether proposition $\psi$ holds. The loan approval task is formulated as a binary classification, where the input space $\mathcal{X} = \mathcal{X}_1 \times \dots \times \mathcal{X}_D \subseteq \mathbb{R}^D$ represents $D$-dimensional applications, and the class label $\mathcal{Y} = \{-1, +1\}$ denotes credit decision, i.e., \{rejected, accepted\}. 
Each application is a vector $x = (x_1, \dots, x_D) \in \mathcal{X}$, and $\mathcal{H}: \mathcal{X} \to \mathcal{Y}$ is the classifier’s decision.
For each rejection, i.e., $\mathcal{H}(\bar{x}) = -1$, the objective is to find an optimal action, represented by a pertubation vector $a = (a_1,..., a_D)\in \mathcal{A}$, such that $\mathcal{H}(\bar{x}+a) = +1$ while minimizing $\mathcal{C}(\bar{x}, \bar{x}+a)$ as the cost of changing from $\bar{x}$ to $\bar{x}+a$. The action space, $\mathcal{A} \subseteq \mathbb{R}^D$ is then defined as a finite $D-$demensional discrete set to ensure the actionability of the explanations.
\label{sect:notes-and-probsetting}
\subsection{Problem Formulation}
The DACE modeling problem is decomposed into 3 independent components: the action space $\mathcal{A}$, the cost function (our work focuses primarily on), and the validity constraint evaluated by the given classifier on the pertubated vector.

\textbf{The cost function}. 
DACE proposed a promising novel cost function, as it combines two efficient distribution-aware metrics: the \textit{Mahalanobis distance} (MD) \cite{de2000mahalanobis} and the LOF \cite{breunig2000lof}. In particular, MD is widely used in statistics to measure distance between two instances wrt. the correlation among features from their data distribution \cite{de2000mahalanobis}. Given two vectors $x, x^\prime \in \mathbb{R}^D$, the formulation of this metric, noted as $d_M(x, x^\prime | \Sigma^{-1}): \mathbb{R}^D\times \mathbb{R}^D \rightarrow \mathbb{R}_{\geq0}$, where $\Sigma^{-1}$ is the inverse covariance matrix of data distribution, is given by:
\begin{equation}
    d_M(x, x^\prime|\Sigma^{-1}) := \sqrt{(x^\prime-x)^\intercal \Sigma^{-1}(x^\prime-x)}
    \label{eq:DMformula}
\end{equation}
As $U\in R^{D\times D}$ is the Cholesky factor of $\Sigma^{-1}$, such that $\Sigma^{-1} = U^\intercal U$, 
we have an $l_2$-normed form of the MD as: $d_M(x,x^\prime|\Sigma^{-1})=||U(x^\prime-x)||_2$.

On the other hand, the LOF (or k-LOF) scores how likely an instance is to be an outlier, compared by its local density to its $k$ nearest neighbors \cite{breunig2000lof}. First, we consider a subset of $N$ instances from the data, $X \subseteq \mathcal{X}$, and a distance metric $\Delta:\mathcal{X}\times \mathcal{X}\rightarrow \mathbb{R}_{\geq 0}$, with the assumption of decomposability: $\Delta(x, x^\prime) = \sum_{d=1}^D{\Delta_d(x_d, x^\prime_d)}, \forall x, x^\prime \in \mathcal{X}$. Next, for each instance $x\in X$, we define $N_k(x)$ as the set of its $k$ nearest neighbors compared by $\Delta$, and $d_k(x)$ measures the distance $\Delta$ between $x$ and its furthest neighbor in $N_k(x)$. Based on these definitions, an asymmetric distance called reachability-distance ($rd_k$) between two instance $x, x^\prime\in X$ is given by: $rd_k(x,x^\prime)= \max\{\Delta(x,x^\prime), d_k(x^\prime)\}$. Therefore, a density-based score for instance $x$ was introduced, denoted by $lrd_k(x)$, which is the inverse of the average reachability distance of $x$ to each $x^\prime$ from its $N_k(x)$ \cite{breunig2000lof}. To this end, LOF of $x$ (k-LOF) is defined as:
\begin{equation}
    q_k(x|X) := \frac{1}{|N_k(x)|} \times \displaystyle \sum_{x^\prime \in N_k(x)}{\frac{lrd_k(x^\prime)}{lrd_k(x)}}
    \label{eq:k-LOF}
\end{equation}

The novel cost of DACE consisted of the squared \textit{Mahalanobis distance} ($d_M^2(\bar{x},\bar{x}+a|\Sigma^{-1})$) and the k-LOF of $\bar{x}+a$ ($q_k(\bar{x}+a|X)$). These two metrics are combined through a trade-off parameter $\lambda \geq 0$, which converts the problem into a single-objective optimization problem \cite{kanamori2020dace}.
However, linearizing this novel cost function into a MILP formulation requires a substantial number of auxiliary variables and constraints, which proved computationally infeasible in their preliminary experiments. A surrogate cost function was used as follows:
\begin{equation}
    \hat{C}_{DACE} = \hat{d}_M(\bar{x}, \bar{x}+a|\Sigma^{-1}) + \lambda.q_1(\bar{x}+a|X)
    \label{eq:surrogated_cost}
\end{equation}

The surrogate function replaces the \textit{Mahalanobis distance} with an $l_1$-norm based distance, which yields a linear form of $\hat{d}_M(\bar{x},\bar{x}+a|\Sigma^{-1}) = ||U.a||_1 = \displaystyle\sum_{d=1}^D{|\langle U_d, a \rangle|}$; and adopts a fixed score of $k=1$ for the LOF, which leads to the simplified formulation:
\begin{equation}
    q_1(\bar{x}+a|X) = lrd_1(x^{(n^*)}).rd_1(\bar{x} +a, x^{(n^*)}) 
    \label{eq:1_lof}
\end{equation} where $x^{(n^*)}$ denotes the nearest neighbor of $\bar{x}+a$ in $X$.

\textbf{The MILP formulation of $\hat{C}_{DACE}$}.
The MILP formulation for the objective function of DACE and its constraints can be written as follows: 

\begin{align}
\text{Minimize}\quad & \sum_{d=1}^D \delta_d 
   + \lambda \sum_{n=1}^N lrd_1(x^{(n)}) .\rho_n \nonumber \\
\text{subject to}\quad 
& \sum_{i=1}^{I_d} \pi_{d,i} = 1, && \forall d \in [D] \label{eq:action_indicator} \\
& - \delta_d \leq \sum_{d^\prime=1}^D \Bigg(U_{d, d^\prime} \sum_{i=1}^{I_{d^\prime}} a_{d^\prime, i}.\pi_{d^\prime, i}\Bigg) \leq \delta_d, 
   && \forall d \in [D] \label{eq:MD_constraint} \\
& \sum_{n=1}^N{\mu_n} = 1, && \forall n \in [N] \label{eq:nearest_indicator} \\
&  \sum_{d=1}^D{\sum_{i=1}^{I_d}{(c_{d,i}^{(n)} - c_{d,i}^{(m)}).\pi_{d, i}}} \leq C_n.(1-\mu_n), 
   && \forall n, m \in [N] \label{eq:nearest_constraint} \\
& \rho_n \geq d_1(x^{(n)}).\mu_n, && \forall n \in [N] \label{eq:reachability_bound} \\
& \rho_n \geq \left( \sum_{d=1}^D{\sum_{i=1}^{I_d}{c_{d,i}^{(n)}}.\pi_{d, i}} \right) - C_n.(1-\mu_n), && \forall n \in [N] \label{eq:reachability_match_nearest}
\end{align}

In this formulation, binary variables are introduced, such as $\pi_{d,i}=\mathbb{I}[a_d = a_{d,i}]$, where $a_{d,i}$ is the $i$-th element of the finite set $A_d$, and $\mu_n=\mathbb{I}[x^{(n)}\in N_1(\bar{x}+a)]$, which indicates whether $x^{(n)} \in X$ is the nearest neighbor of $\bar{x}+a$. Continuous variables include $\delta_d$, the $d$-axis statistical distance between $\bar{x}$ and $\bar{x}+a$, and $\rho_n$, the reachability distance from $x^{(n)}$ to $\bar{x}+a$ if $x^{(n)} \in N_1(\bar{x}+a)$ \cite{kanamori2020dace}. The constraint \eqref{eq:MD_constraint} defines the $l_1$-based Mahalanobis distance, while constraints from \eqref{eq:nearest_indicator} to \eqref{eq:reachability_match_nearest} specify the 1-LOF term. Within the discrete action space $\mathcal{A}$ and given subset $X$, DACE further employs constants $c_{d,i}^{(n)}:=\Delta_d(\bar{x}_d +a_{d,i}, x_d^{(n)})$ and $C_n = \max_{a\in \mathcal{A}} \Delta(\bar{x}+a, x^{(n)})$.
As analyzed in DACE, the MILP formulation of $\hat{C}_{DACE}$ requires $\mathcal{O}(D+N^2)$ auxiliary constraints: $\mathcal{O}(D)$ from the Mahalanobis-based term and $\mathcal{O}(N^2)$ from the 1-LOF. The main source of complexity is constraint \eqref{eq:nearest_constraint}, which introduces $N^2$ inequalities.

\label{subsubsec:cost_milp}

\section{Proposed Methods}
\label{sec:method}
Although a surrogate cost function (Eq. \ref{eq:surrogated_cost}) is used, the undesired $\mathcal{O}(N^2)$ constraints from the 1-LOF term impose limitations on both research and the extension of DACE to real datasets. Our first contribution addresses this issue by introducing an alternative formulation for $\mu_n = \mathbb{I}[x^{(n)} \in N_1(\Bar{x}+a)]$. We then consider extending DACE to a linear SVM under standard scaling, which is a common preprocessing step.
\subsection{Reducing constraints for 1-LOF}
\label{subsec:reduce-LOF}
A principal objective is to find the instance $x^{(n)}\in X$, where $X$ contains $N$ samples, that minimizes $\Delta(\bar{x}+a, x^{(n)})$. To this end, we define the finite set $T=\{\Delta(\bar{x}+a, x^{(1)}), ..., \Delta(\bar{x}+a, x^{(N)})\}$ and introduce a continuous variable $t\in \mathbb{R}$ to represent its minimum value. Inspired by the integer programming formulation of maximum constraints in \cite{aps2020mosek}, we propose a minimization counterpart for our problem as follows:
\begin{equation}
\begin{cases}
    & \Delta(\bar{x}+a, x^{(n)}) - M.(1-\mu_n) \leq t \leq \Delta(\bar{x}+a, x^{(n)}) \quad \forall n \in [N] \label{eq:proposed_idea}\\
    & \mu_n \in \{0; 1\} \quad \forall n \in  [N] \\
    & \displaystyle \sum_{n=1}^N{\mu_n} = 1
\end{cases}
\end{equation}

We utilize the indicator $\mu_n$ as the DACE, but now a new big-M bounding approach is used to replace constraint \eqref{eq:nearest_constraint}. These $2N$ constraints consist of $N$ ``less-than-or-equal-to'' inequalities, which iteratively tighten the upper bound (UB) of $t$ until $t \leq min(T)$, and $N$ 'greater-than-or-equal-to' inequalities, which provide the corresponding lower bound (LB) for $t$. Notably, the lower bound constraints incorporate the indicator $\mu_n$ through a large constant value M, ensuring that when $\mu_n=1$, we have $t = \Delta(\bar{x}+a, x^{(n)})$. Together with the tightest upper bound of $t$, it follows that $t = \Delta(\bar{x}+a, x^{(n^*)}) = min(T)$ if and only if $\mu_{n^*} = 1$.

Even though the constant $M$ must be sufficiently large, we suggest a lower bound given by $M \geq \displaystyle\max_{n\in[N]}{C_n}$. This is not the tightest bound for $M$ to ensure the feasibility of the model, but this approach utilizes the pre-computed constants $C_n$, which are also used to restrict the $\rho_n$ in \eqref{eq:reachability_match_nearest}.

Note that the metric $\Delta$ is decomposable, and with the defined $c_{d,i}^{(n)}$, it holds that $\sum_{d=1}^D\sum_{i=1}^{I_d}c_{d,i}^{(n)}.\pi_{d,i} = \Delta(\bar{x}+a, x^{(n)})$.
The following MILP formulation is proposed as a replacement for the $N^2$ inequalities in \eqref{eq:nearest_constraint}:
\begin{equation}
    \sum_{d=1}^D\sum_{i=1}^{I_d}{c_{d,i}^{(n)}.\pi_{d,i}} - M.(1-\mu_n) \leq t \leq \sum_{d=1}^D\sum_{i=1}^{I_d}{c_{d,i}^{(n)}.\pi_{d,i}} \quad \forall n \in [N]
    \label{eq:N1_proposed_milp}
\end{equation}

To summarize, while the value of $t$ is not intended to be used in any further computational step, our method to model the $N_1(\bar{x}+a)$ shows an efficient reduction in auxiliary constraints, from $N^2$ to $2N$, which is supposed to make the MILP formulation cleaner and more computational efficient.

\subsection{Extending DACE to Linear SVM with standard scaler}

Given a well-trained linear SVM model with weight vector $w^\star$ and intercept $b^\star$, learned from the training data $\mathcal{X}^{train} \subset \mathcal{X}$. The prediction for a new instance $x \notin \mathcal{X}^{train}$ is defined as $\mathcal{H}(x) = \operatorname{sign}(\langle w^\star, x \rangle + b^\star)$, where the sign function is defined as $\operatorname{sign}(z) = +1$ if $z \geq 0$ and $\operatorname{sign}(z) = -1$ otherwise. 

When using a standard scaler in preprocessing, the decision function is not applied directly to the original $x$, but to the scaled instance $x^{scaled}$. Note that $\mathcal{X}^{train} \subset R^D$, we only apply the standard scaler to the subset of feature indices $D^\prime \subseteq [D]$, corresponding to the non-binary features.. For each $d \in D^\prime$, we denote $\bar{\mathcal{X}}_d$ and $\sigma_d$ as the mean and standard deviation of the feature $x_d$, computed from its values in $\mathcal{X}_d^{train}$. For $d\in [D] \setminus D^\prime$, we assume that $\bar{\mathcal{X}}_d = 0$ and $\sigma_d=1$. Then, we have the unified transformation: $\quad x_d^{scaled} = \displaystyle\frac{x_d - \bar{\mathcal{X}}_d}{\sigma_d}, \quad \forall d\in [D]$.

 Regarding CE (Sect. \ref{sect:notes-and-probsetting}), the validity constraint for $\mathcal{H}(\bar{x}+a)=+1$ is that $\langle w^\star, (\bar{x}+a)^{scaled} \rangle + b^\star \geq 0$. Since the action space $\mathcal{A}$ is constructed from the original training data (feature values are unscaled), the integration problem is to express $(\bar{x}+a)^{scaled}$ as a linear function of $\bar{x}^{scaled}$ (constant) and $a$. We have:
\begin{equation}
    (\bar{x}+a)_d^{scaled} = \frac{(\bar{x}_d+a_d) - \bar{\mathcal{X}}_d}{\sigma_d} = \bar{x}_d^{scaled} + \frac{a_d}{\sigma_d} \quad \forall d\in [D]
    \label{eq:scaled_x+a}
\end{equation}

The MILP formulation for the $\langle w^\star, (\bar{x}+a)^{scaled} \rangle + b^\star \geq 0$ is given by:
\begin{equation}
    \sum_{d=1}^D{\sum_{i=1}^{I_d}{\frac{w_d^\star.a_{d,i}}{\sigma_d}.\pi_{d,i}}} + \left(\sum_{d=1}^D{w_d^\star.\bar{x}_d^{scaled} } + b^\star \right) \geq 0
    \label{eq:scaledSVM_milp}
\end{equation}

Because the MILP formulations of the cost function and the validity constraints imposed by the classifier are independent components, the constraint \eqref{eq:scaledSVM_milp} can be directly appended to the formulation in Sect. \ref{subsubsec:cost_milp}, yielding an adapted MILP formulation of DACE that is ready to be solved by efficient solvers such as IBM ILOG CPLEX\footnote{https://www.ibm.com/products/ilog-cplex-optimization-studio}.

\section{Evaluation}
\label{sec:evaluation}
In this section, we evaluate the performance of the proposed method in terms of the quality of the CE and the effectiveness of the new formulation for 1-LOF.
\subsection{Datasets}
\label{subsec:data}
We evaluate our proposed method on two well-known credit approval datasets: German Credit\footnote{https://archive.ics.uci.edu/dataset/144/statlog+german+credit+data} and Home Equity Line of Credit (HELOC\footnote{https://www.kaggle.com/datasets/averkiyoliabev/home-equity-line-of-creditheloc}). 
The German Credit dataset consists of 1,000 records, represented by 13 categorical and 7 numerical attributes, with no missing values. We use one-hot encoding to transform the categorical attributes. The task is to determine the potential risk of issuing credit to an applicant.  
The HELOC dataset is provided by FICO corporation\footnote{https://www.fico.com/}, contains 10,459 records described by 23 numerical attributes of credit line applications, with a missing-values rate of 5.6\%, which is removed during preprocessing. The main objective is to predict whether they will successfully repay their HELOC account within a two-year period. After pre-processing, the class label proportions are 70\% ``Good'' compared to 30\% ``Bad'' for German Credit, and 48\% and 52\% for ``Good'' and ``Bad'', respectively, for HELOC.

\subsection{Experimental setups}
\label{subsec:settings}
\begin{figure}[h]
\centering
\includegraphics[width=1\linewidth]{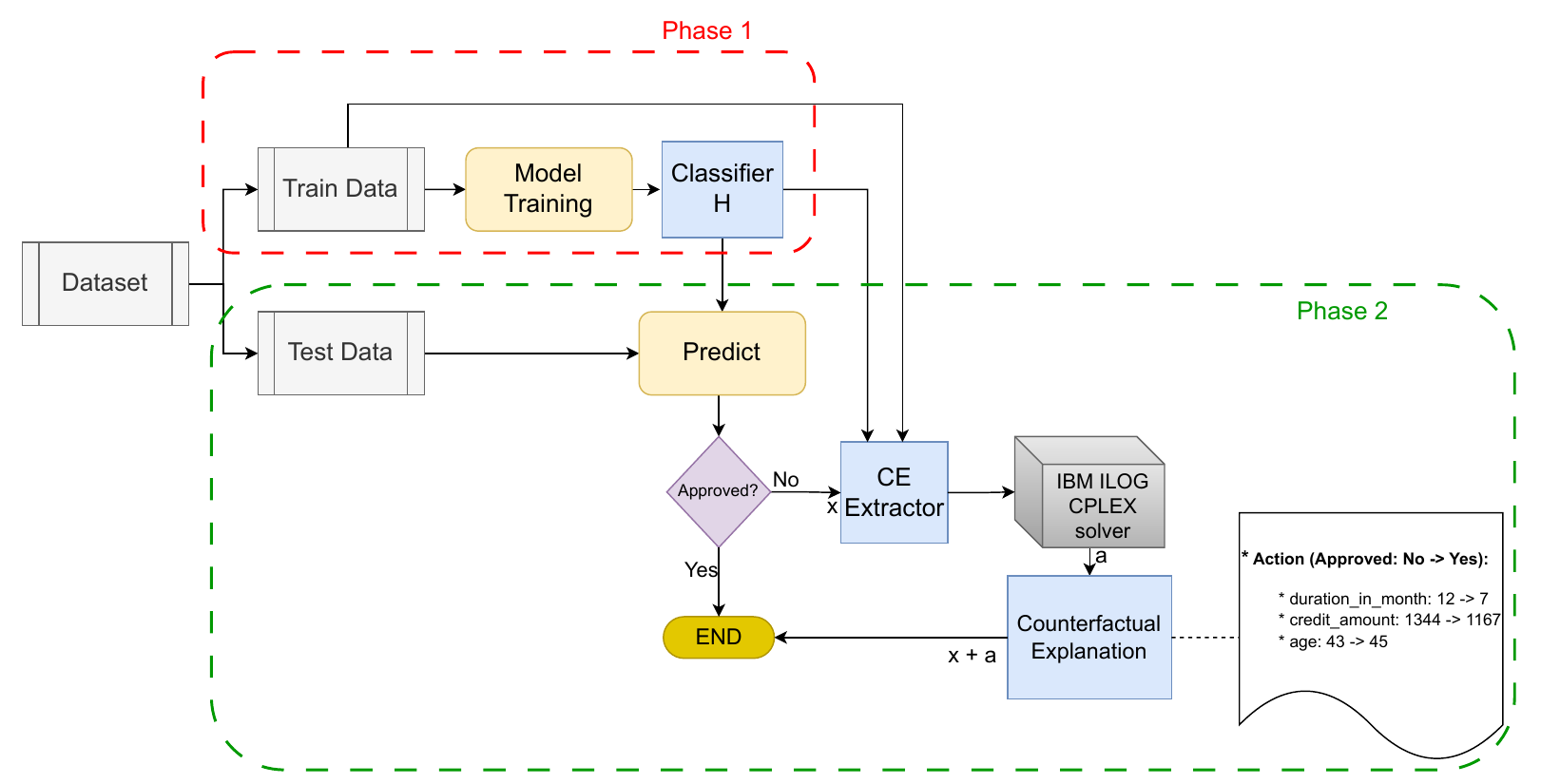}
\caption{Overview of the experimental process}
\label{fig:pipeline}
\vspace{-10pt}
\end{figure}

The experimental process includes two phases as illustrated in Fig. \ref{fig:pipeline}. In which, the CE search problem is formulated as a MILP by the CE Extractor component, and the IBM ILOG CPLEX solver is employed to find the optimal solution. The dataset is split into training and test sets using a 75:25 ratio with the single-split hold-out method for classifier validation.
For each dataset and classifier $\mathcal{H}$, counterfactual explanations are generated for the 10 rejected test instances. The experiments are executed on Intel(R) Core(TM) i5-10500H CPU 6x12@2.50GHz, 8GB RAM; IBM ILOG CPLEX solver v22.1.2.

\textbf{Hyperparameter settings}: We follow the settings in the baseline \cite{kanamori2020dace} for logistic regression (LR) and random forest (RF) classifiers: logistic regression with L2-regularization ($C = 1.0$), and random forest with 100 trees and a maximum depth of 6. For the linear SVM, input attributes are standardized using a standard scaler with the soft-margin penalty parameter $C = 1.0$. With respect to the MILP problem, the maximum number of attributes allowed to change is set to 4; the coefficient of the 1-LOF term in the optimization objective is $\lambda = 0.01$ for the German Credit dataset and $\lambda = 1.0$ for the HELOC dataset. To ensure consistency with the experiments on logistic regression and random forest classifiers in \cite{kanamori2020dace}, $N$ is fixed at 20; however, for the German Credit and HELOC datasets, $N$ is varied incrementally with $N \in \{20, 50, 100, 200\}$. The time limit for the CPLEX solver is set to 1200 seconds for $N \in \{20, 50\}$ and 3600 seconds for $N \in \{100, 200\}$.

\subsection{Experimental Results}
\label{subsec:results}
The performance of three classifiers on the two datasets are presented in Table \ref{tbl:performance_classifier}. The best values are in \textbf{bold}. In the German Credit dataset, LR shows the best accuracy (73.76\%) and precision (78.57\%) while RF has the best performance on the recall (97.71\%) and F1-score (83.68\%), which indicates that the RF model is affected by the class imbalance issue of this dataset. An opposite result is observed in the HELOC dataset, where the linear SVM outperforms the LR and RF models in terms of accuracy, recall, and F1-score. This can be attributed to the effect of the standardization step using the standard scaler.

\begin{table}[!h]
\vspace{-15pt}
\centering
\caption{Performance of classification models}
\label{tbl:performance_classifier}
\begin{tabular}{|l|c|c|c|c|}
\hline
\multicolumn{5}{|c|}{\textbf{German Credit}} \\ \hline
\textbf{Models} & \textbf{Accuracy} (\%) & \textbf{Precision} (\%) & \textbf{Recall} (\%) & \textbf{F1-score} (\%) \\
\hline
LR  & $\mathbf{73.76 \pm 2.60}$ & $\mathbf{78.57 \pm 1.90}$ & $85.63 \pm 3.86$ & $81.91 \pm 2.05$ \\
RF  & $73.52 \pm 1.24$ & $73.19 \pm 1.09$ & $\mathbf{97.71 \pm 1.02}$ & $\mathbf{83.68 \pm 0.81}$ \\
SVM & $72.88 \pm 2.93$ & $77.93 \pm 2.21$ & $85.16 \pm 3.26$ & $81.34 \pm 2.16$ \\
\hline
\multicolumn{5}{|c|}{\textbf{HELOC}} \\ \hline

LR  & $73.18 \pm 0.97$ & $72.23 \pm 1.49$ & $71.04 \pm 1.77$ & $71.63 \pm 1.25$ \\
RF  & $73.11 \pm 0.99$ & $\mathbf{73.02 \pm 1.23}$ & $69.13 \pm 1.40$ & $70.82 \pm 1.21$ \\
SVM & $\mathbf{73.36 \pm 0.98}$ & $72.36 \pm 1.59$ & $\mathbf{71.36 \pm 1.35}$ & $\mathbf{71.86 \pm 1.29}$ \\
\hline
\end{tabular}
\vspace{-10pt}
\end{table}

\begin{figure*}[!h]
\vspace{-10pt}
\centering
\begin{subfigure}{.49\linewidth}
    \centering
    \includegraphics[width=\linewidth]{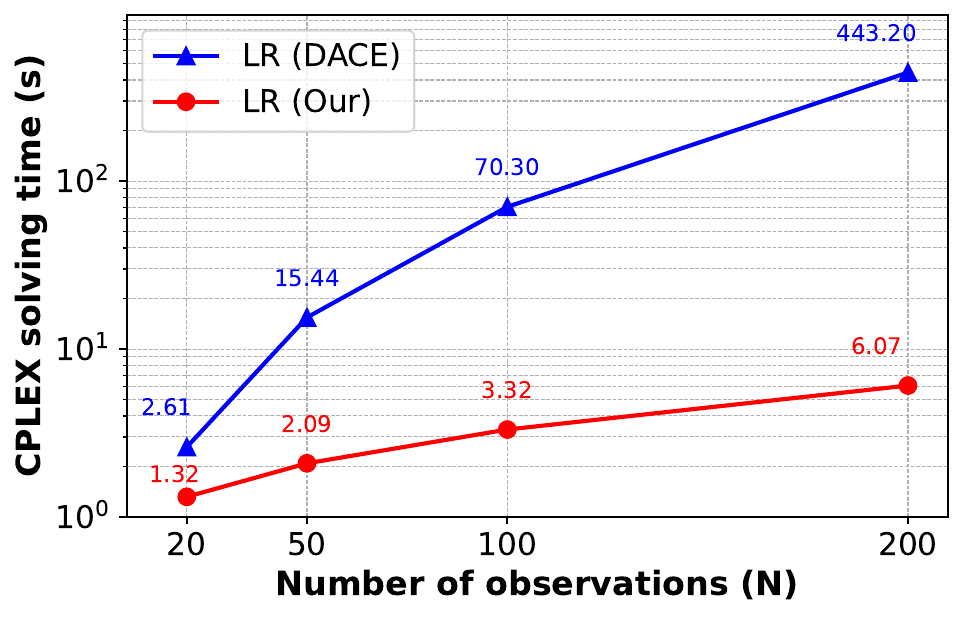}
    \caption{German Credit}
\end{subfigure}
\hfill
\begin{subfigure}{.49\linewidth}
    \centering
    \includegraphics[width=\linewidth]{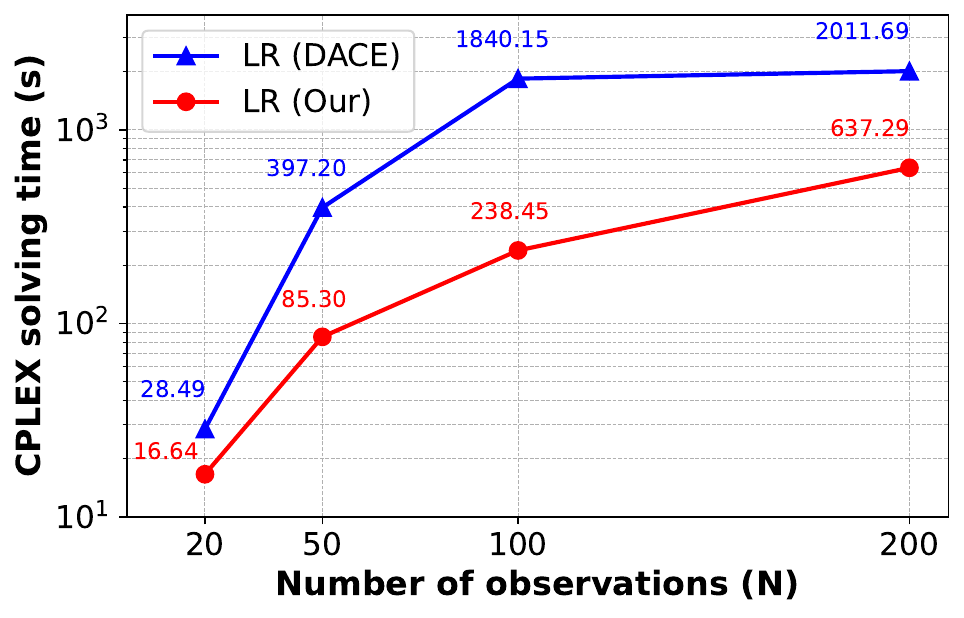}
    \caption{HELOC}
\end{subfigure}    
\hfill
\vspace{-5pt}
\caption{Running time for solving the MILP problem (LR classifier)}
\label{fig:running_time_lr}
\vspace{-10pt}
\end{figure*}

\begin{figure*}[!h]
\vspace{-10pt}
\centering
\begin{subfigure}{.49\linewidth}
    \centering
    \includegraphics[width=\linewidth]{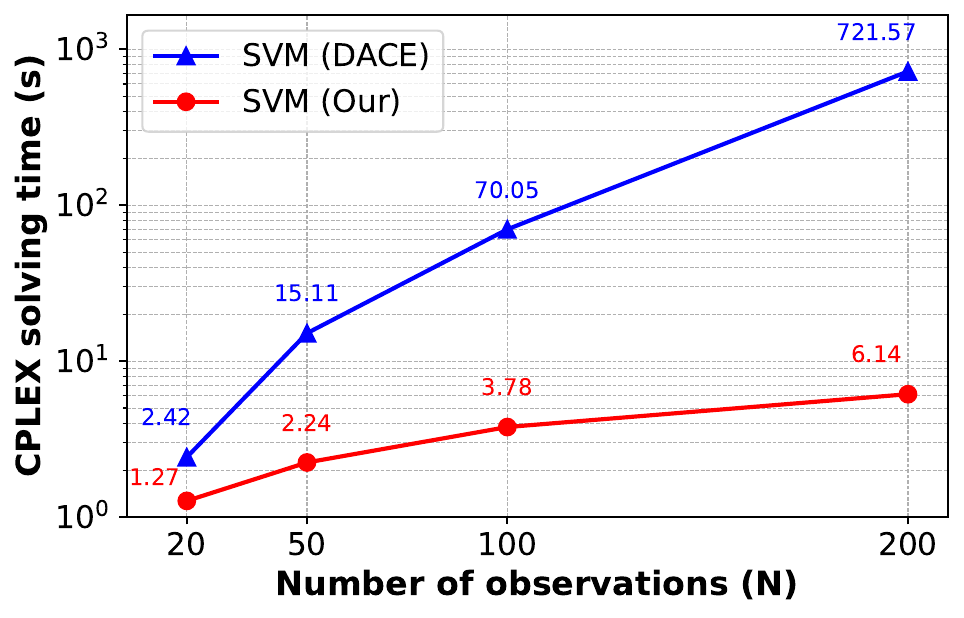}
    \caption{German Credit}
\end{subfigure}
\hfill
\begin{subfigure}{.49\linewidth}
    \centering
    \includegraphics[width=\linewidth]{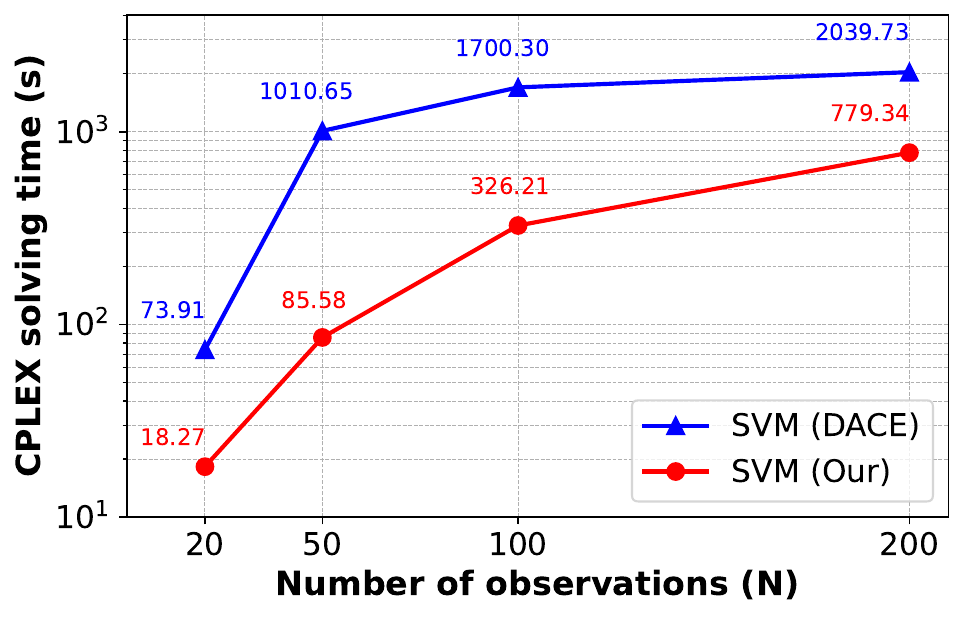}
    \caption{HELOC}
\end{subfigure}    
\hfill
\vspace{-5pt}
\caption{Running time for solving the MILP problem (linear SVM classifier)}
\label{fig:running_time_svm}
\vspace{-10pt}
\end{figure*}

In the next experiment, we evaluate the performance of IBM ILOG CPLEX solver on Mahalanobis distance (MD) (Eq. \ref{eq:DMformula}), k-LOF index (with $k=10$, i.e., 10-LOF) (Eq. \ref{eq:k-LOF}) (smaller is better) and the actual running time to solve the MILP problem regarding counterfactual explanations and our proposed method to reduce the number of constraints for 1-LOF from $\mathcal{O}(N^2)$ to $\mathcal{O}(N)$. It is observed that the time required to solve the optimization problem decreased by at least 50\% when the number of observations is $N = 20$ (Table \ref{tbl:result_MILP}). Simultaneously, the solutions preserved optimal quality, with the same MD and low 10-LOF values, confirming coherence with the data distribution and non-outlier behavior, as in the original DACE model. Furthermore, our proposed 1-LOF formulation significantly reduces the number of constraints, allowing larger values of $N$ without a substantial increase in computational cost. As illustrated in Fig. \ref{fig:running_time_lr} and Fig. \ref{fig:running_time_svm}, our method (red line) outperforms the DACE method (blue line), as the time required to solve the MILP problem is significantly reduced.

\begin{table}[h]
\vspace{-15pt}
\centering
\caption{Counterfactual explanation search ($N=20$) }
\label{tbl:result_MILP}
\begin{tabular}{|l|c|c|c|c|}
\hline
\multicolumn{5}{|c|}{\textbf{German Credit}} \\ \hline
\textbf{Models} & \textbf{MD} & \textbf{10-LOF} & \textbf{Time (s) - DACE} & \textbf{Time (s) - Our} \\
\hline
LR  & 1.90 $\pm$ 1.29 & 1.04 $\pm$ 0.07 & 2.61 $\pm$ 0.50   & \textbf{1.32 $\pm$ 0.25} \\
RF  & \textbf{1.12 $\pm$ 1.09} & \textbf{0.97 $\pm$ 0.03} & 59.73 $\pm$ 27.53 & \textbf{35.75 $\pm$ 31.98} \\ 
SVM & 2.75 $\pm$ 0.99 & 0.98 $\pm$ 0.07 & \textbf{2.42 $\pm$ 0.43}   & \textbf{1.27 $\pm$ 0.29} \\
\hline
\multicolumn{5}{|c|}{\textbf{HELOC}} \\ \hline
LR  & 3.38 $\pm$ 1.30 & 1.08 $\pm$ 0.09 & \textbf{28.49 $\pm$ 8.42}    & \textbf{16.64 $\pm$ 8.04} \\
RF  & 3.14 $\pm$ 1.02 & 1.12 $\pm$ 0.10 & 1114.90 $\pm$ 192.00 & \textbf{242.76 $\pm$ 80.48} \\ 
SVM & \textbf{3.07 $\pm$ 1.14} & \textbf{1.07 $\pm$ 0.09} & 73.91 $\pm$ 46.84   & \textbf{18.27 $\pm$ 9.08} \\
\hline
\end{tabular}
\vspace{-15pt}
\end{table}

\section{Conclusions and Outlook}
\label{sec:conclusion}
\vspace{-5pt}
In this work, we first propose a reformulation for the 1-LOF component from DACE, which reduces the complexity of model. Then, we extend the DACE model to a linear SVM classifier with an input standardization phase. Experiments on real-world datasets show improvements in running time to find optimal explanation. 
In the future, the model can be extended to more complex classifiers such as boosting trees, stacking models, or deep neural networks. Remaining challenges include the increasing number of weak learners or perceptrons in modern classifiers, and the auxiliary cost of modeling non-linear activation functions such as the sigmoid. 
In addition, the quality of explanations may be enhanced by incorporating additional criteria such as causality or behavioral constraints.
%
%
%
\vspace{-5pt}
\bibliographystyle{splncs04}
\bibliography{bibliography}

\end{document}